\documentclass[conference]{IEEEtran}
\IEEEoverridecommandlockouts

\usepackage{cite}
\usepackage{amsmath,amssymb,amsfonts}
\usepackage{algorithmic}
\usepackage{graphicx}
\usepackage{textcomp}
\usepackage{xcolor}
\usepackage{enumitem} 
\usepackage{url}

\usepackage{kotex}
\usepackage{times}
\usepackage{soul}
\usepackage{url}
\usepackage[hidelinks]{hyperref}
\usepackage[utf8]{inputenc}
\usepackage[small]{caption}
\usepackage{graphicx}
\usepackage{amsmath}
\usepackage{amsthm}
\usepackage{booktabs}
\usepackage[switch]{lineno}
\usepackage{multirow}

\usepackage{amsmath,amssymb,amsfonts}
\usepackage{graphicx}
\usepackage{textcomp}
\usepackage{xcolor}
\usepackage{enumitem} 
\usepackage[linesnumbered,ruled,vlined]{algorithm2e}
\usepackage{float}
\usepackage{dblfloatfix}
\usepackage{makecell} 
\usepackage{enumitem}

\usepackage{array}
\usepackage{placeins}
\usepackage[linesnumbered,ruled]{algorithm2e}

\def\BibTeX{{\rm B\kern-.05em{\sc i\kern-.025em b}\kern-.08em
    T\kern-.1667em\lower.7ex\hbox{E}\kern-.125emX}}
\begin{document}

\title{Target Circuit Matching in Large-Scale Netlists Using GNN-Based Region Prediction \\
}

\vspace{-10mm}

\author{\IEEEauthorblockN{Sangwoo Seo$^1$, Jimin Seo$^1$, Yoonho Lee$^1$, Donghyeon Kim$^{2}$, Hyejin Shin$^{2}$, Banghyun Sung$^{2}$, Chanyoung Park$^{1*}$\thanks{*Corresponding Author}}
\IEEEauthorblockA{\textit{KAIST, Republic of Korea$^{1}$}, \textit{SK hynix, Republic of Korea$^{2}$}\\
\{sangwooseo, jimin.seo, sml0399benbm, cy.park\}@kaist.ac.kr, \{donghyeon6.kim, hyejin2.shin, banghyun.sung\}@sk.com}








}
\vspace{-10mm}
\maketitle

\vspace{-10mm}

\begin{abstract}
Subgraph matching plays an important role in electronic design automation (EDA) and circuit verification. 
Traditional rule-based methods have limitations in generalizing to arbitrary target circuits. Furthermore, node-to-node matching approaches tend to be computationally inefficient, particularly for large-scale circuits.
Deep learning methods have emerged as a potential solution to address these challenges, but existing models fail to efficiently capture global subgraph embeddings or rely on inefficient matching matrices, which limits their effectiveness for large circuits.
In this paper, we propose an efficient graph matching approach that utilizes Graph Neural Networks (GNNs) to predict regions of high probability for containing the target circuit.
Specifically, we construct various negative samples to enable GNNs to accurately learn the presence of target circuits and develop an approach to directly extracting subgraph embeddings from the entire circuit, which captures global subgraph information and addresses the inefficiency of applying GNNs to all candidate subgraphs.
Extensive experiments demonstrate that our approach significantly outperforms existing methods in terms of time efficiency and target region prediction, offering a scalable and effective solution for subgraph matching in large-scale circuits. 
The source code is available at \url{https://github.com/sang-woo-seo/Circuit-matching-GNN}.
\end{abstract}


\begin{table*}[!b]
\centering
\vspace{-2mm}
\caption{Node Types and Edge Types in Circuit Graph}
\vspace{-2mm}
\renewcommand{\arraystretch}{0.8} 
\begin{tabular}{c|cccccccc}
\hline
          & 0   & 1   & 2   & 3    & 4    & 5         & 6        & 7        \\ \hline
Node type & vdd & gnd & net & PMOS & NMOS & capacitor & resistor & inductor \\ \hline
\end{tabular}

\vspace{1.5mm} 

\renewcommand{\arraystretch}{1.4} 
\resizebox{\textwidth}{!}{%
\begin{tabular}{c|cccccccccccccc}
\hline
          & 0          & 1         & 2           & 3         & 4          & 5         & 6           & 7         & 8  & 9  & 10 & 11 & 12 & 13 \\ \hline
Edge type & PMOS-drain & PMOS-gate & PMOS-source & PMOS-base & NMOS-drain & NMOS-gate & NMOS-source & NMOS-base & C+ & C- & R+ & R- & I+ & I- \\ \hline
\end{tabular}%
}

\label{tab:node/edge types}
\end{table*}

\section{Introduction}
Subgraph matching plays a critical role in generating hierarchical netlists and verifying circuits, making it a valuable technique in electronic design automation (EDA) and circuit verification.
Circuit verification determines whether the target circuit exists within the entire circuit when faults are detected or components require replacement.
For example, it identifies fault areas that lead to major malfunctions or connections vulnerable to electrostatic discharge (ESD).
Consequently, subgraph matching supports the systematic management of circuit design and ensures efficient circuit verification.

Traditionally, circuit matching relied on a rule-based approach using pre-defined rules. However, this method depends on specific rules for each target circuit, making it unsuitable for application to arbitrary target circuits, as the rules must be redefined whenever the target circuit changes. 
To address this problem, researchers have explored graph-theoretic discoveries \cite{dong2021subcircuit,rajarathnam2020regds} for graph matching, applied graph partitioning to identify potential matching locations \cite{ohlrich1993subgemini}, and formulated the circuit matching problem as a progressive assignment algorithm \cite{rubanov2003subislands}.
However, these methods lead to a significant increase in the time required to match large-scale circuits, as they rely on node-to-node matching between two graphs, such as optimizing the matching matrix.
{VF2 algorithm \cite{cordella2004sub}} uses combinatorial search algorithms for low computational complexity, but its NP-complete nature still imposes limitations on scaling to large circuits. 
These time-consuming inefficiencies make it challenging to achieve immediate responses for circuit matching in large-scale circuits.

With the continuous increase in the scale and complexity of circuits, deep learning techniques have emerged as a key approach to addressing the critical challenge of improving the design efficiency of EDA tools in the circuit design process~\cite{bustany2015ispd,yang2022versatile,dong2023cktgnn}.
Recently, several attempts have been made to apply deep learning to the field of graph matching. However, \cite{bai2019simgnn} focuses on calculating similarity scores without explicitly identifying the matching subgraph, which limits its ability to determine the exact location of the matching subgraph.
\cite{roy2022interpretable}  relies on a matching matrix that scales with the size of the graph, making it unsuitable for large-scale circuits.
Additionally, \cite{lou2020neural} uses order embeddings to represent each node in the embedding space, where subgraph relationships are reflected.
However, by equating the center node embedding with the subgraph embedding, it fails to capture the global information of the subgraph, leading to less accurate predictions of subgraph relationships and reduced effectiveness for matching.



To address these issues, we develop an efficient matching network capable of being applied to large-scale circuits, based on accurate predictions of subgraph relationships.
Our model utilizes Graph Neural Networks (GNN) to predict regions with a high probability of containing the target circuit and prioritizes matching in the ranked regions, enabling efficient subgraph matching in large-scale circuits.
Specifically, our approach addresses several key considerations. 
First, we explore {an effective method} for representing MOSFET-based netlists as graphs to support the GNN training. 
Second, we generate effective negative samples to enable the GNN to accurately learn the presence or absence of target circuits in the entire circuit. In this process, we introduce four types of negative samples including hard negative samples: Partial, Mutation, Others, and Random.
Third, to address the inefficiency of applying the GNN to all candidate subgraphs, we develop a technique to directly extract embeddings of the candidate subgraphs from the entire circuit by utilizing node embeddings from all hops. {This approach specifically focuses on excluding information beyond subgraph boundaries from entire graph.}

{We conducted experiments to evaluate the performance of our model in graph matching.
The experimental results show that our model accurately matches all target circuits within the entire circuit and achieves superior time efficiency compared to other baselines both for matching all target circuits and for matching a single target circuit. 
It is important to note that our method achieves highly efficient matching based on the predicted regions with a high likelihood of containing the target circuit. In our experiments, we also show that our proposed method of target region prediction is superior over various baselines.
}  

In summary, our main contributions are summarized as:
\begin{itemize}[leftmargin=0.5cm]
    \item We propose a method to improve matching efficiency by extracting candidate K-hop subgraphs and identifying regions with a high probability of containing the target circuit using GNN training.
    \item We introduce four types of negative samples, including hard negative samples, to enable the GNN to determine the presence of the target circuit within K-hop subgraphs. 
    \item To improve efficiency, we develop an approach to directly obtaining the K-hop subgraph embeddings from the entire circuit by utilizing node embeddings from all hops.
    \item Extensive experiments demonstrate that our method outperforms various baselines in terms of both time efficiency and target region prediction.
    
\end{itemize}

\section{Related Works} \label{apx:related_works}

\subsection{Subgraph Matching}
Given a target graph and an entire graph, subgraph matching determines whether the entire graph includes the target graph as a subgraph. This task can be performed either by analyzing the graph structure ~\cite{Ullman} or by leveraging node features to establish the relationship between nodes in the graph ~\cite{Coffman,Aleman}. Over time, various methods have been developed to approach this challenge. Some methods rely on similarity search techniques to determine the degree of matching between two graphs~\cite{Yan2005simsearch,Dijkman2009simsearch}, which often involves comparing structural or feature-based representations of the graphs. The similarity between two graphs can be measured either as an exact match~\cite{Shasha2002graphsearching,Yan} or based on their structural similarity~\cite{Willett1998simsearch,Ray2002simsearch}. Since subgraph matching is an NP-complete problem, finding efficient and scalable solutions remains a significant challenge, especially as the size of the graphs increases. Traditional combinatorial search methods such as ~\cite{cordella2004sub,Ullman} are often impractical for large graphs. Consequently, algorithms that can handle subgraph matching within a practical time amount for real-world datasets have been developed. For efficient searching, some approaches utilize graph indexing techniques~\cite{Yan} to filter out irrelevant portions of the data graph. While effective, these methods become increasingly computationally expensive in terms of time and space as the graph size grows. Triangle counting-based methods~\cite{Eden2015countingtraingle,Tsourakakis2008fastcounting} have also been proposed, but they are unsuitable for handling large entire graphs. 
Our method extracts K-hop subgraphs by sampling node embeddings derived from the entire graph. We then estimate the probability of the target graph existing within each extracted subgraph, sort them accordingly, and verify the subregions with the highest probabilities first.

\subsection{Neural Graph Matching}

\cite{Scarselli2009} demonstrated in their work that GNNs can be effectively applied to subgraph matching, highlighting their suitability for addressing problems in relational domains. Graph convolutional networks~\cite{kipf2016semi}, which update nodes through message passing from neighboring nodes, have served as the foundation for GNNs and have been applied across various domains. GNNs are not limited to supervised prediction tasks such as node prediction or edge prediction but are also extended to tasks like measuring the similarity between two graphs, enabling their application to graph isomorphism problems~\cite{bai2019simgnn,li2019graph,lou2020neural,roy2022interpretable}. \cite{bai2019simgnn} calculates the similarity score between two graphs based on the interaction between graph-level embeddings and node-level embeddings. \cite{li2019graph} calculates the similarity score using a cross-graph attention-based mechanism. \cite{lou2020neural} determines whether the target graph is contained within the entire graph using order embeddings. \cite{roy2022interpretable} performs subgraph matching based on edge alignment, leveraging edge-consistent mapping to ensure accurate matching.

\section{Methodology}

\begin{figure}[t] 
\begin{center}
\includegraphics[width=0.9\linewidth]{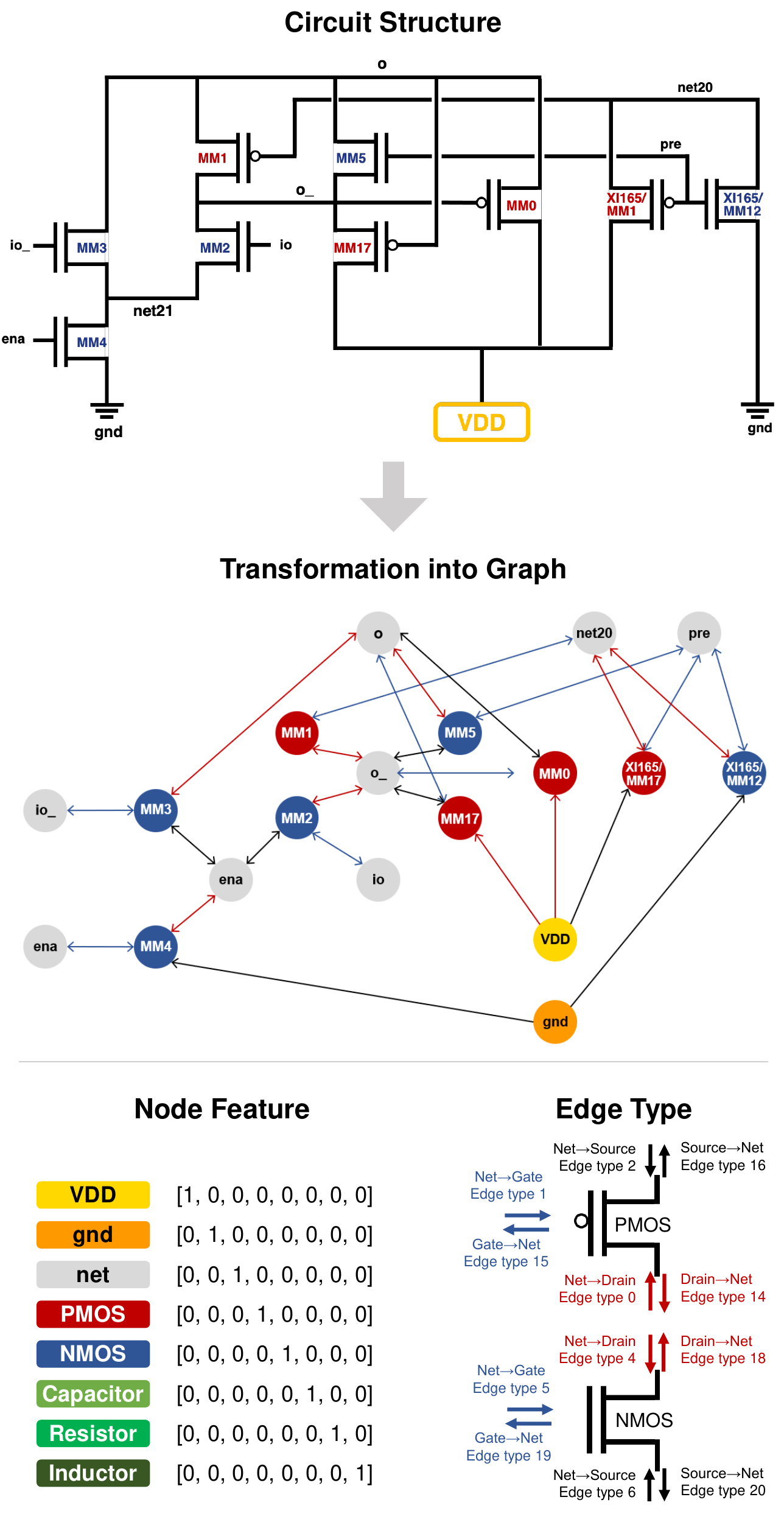}
\end{center}
\vspace{-4mm}
\caption{
Transformation of Circuit Structure into Graph}
\label{fig:Graph_Visulization}
\vspace{-5mm}
\end{figure}

\begin{figure*}[t]
  \centering
  \includegraphics[width=0.97\textwidth]{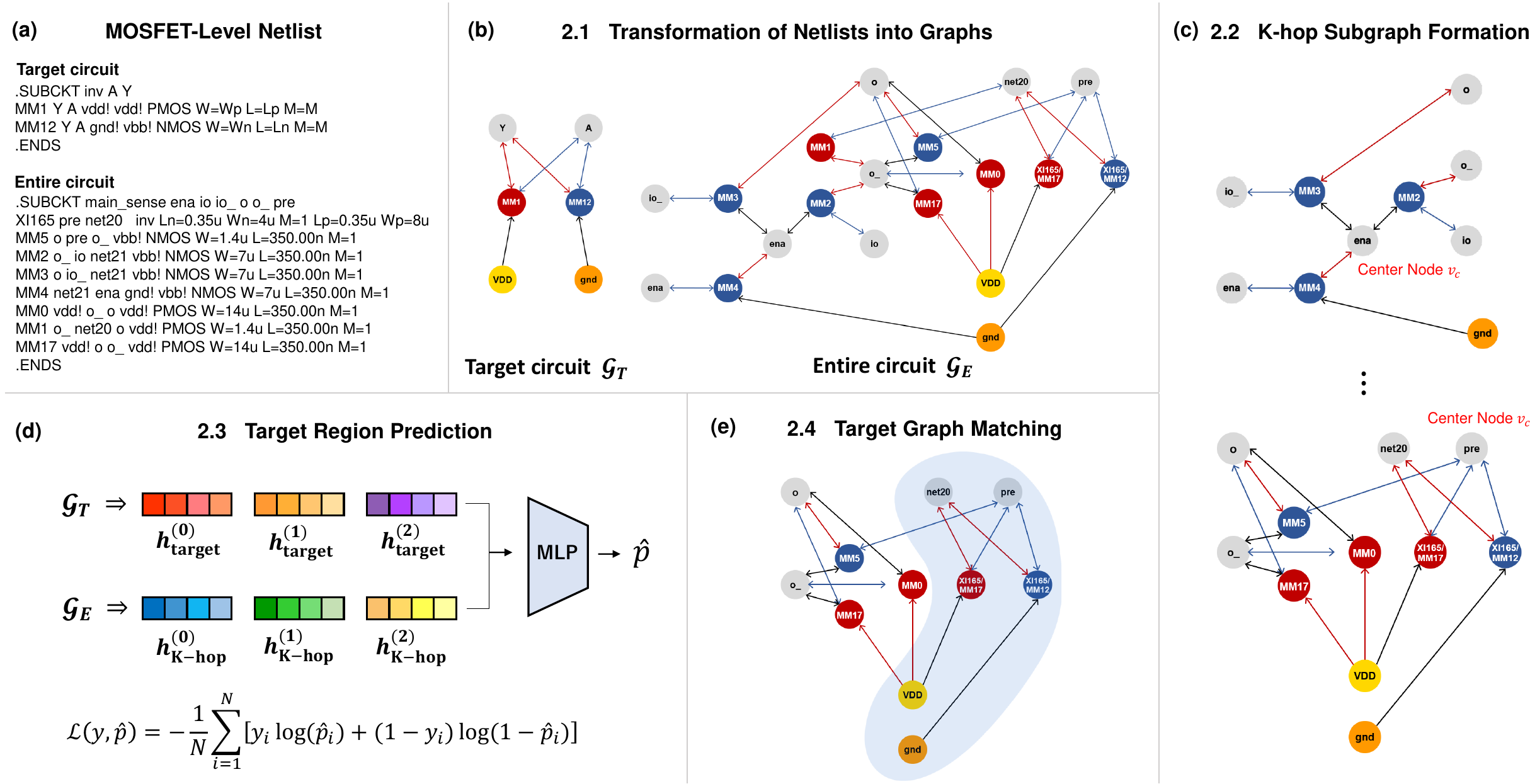}
  \vspace{-1mm}
  \caption{The architecture of our proposed method.} 
  \label{fig:architecture}
  \vspace{-3mm}
\end{figure*}

\subsection{Transformation of Netlists into Graphs}\label{subsec:Netlist to graph}
The circuit design is represented as a netlist composed of cells and nets, where cells represent electronic device units, and nets indicate the connectivity between cells (i.e., hyperedges).
The circuit graph is expressed as 
$\mathcal{G} = (\mathcal{V}, \mathcal{E}, \mathbf{A}, \mathbf{X})$, where $\mathcal{V}$, $\mathcal{E}$, $\mathbf{A}$, and $\mathbf{X}$ denote the set of nodes, edges, the adjacency matrix, and node features, respectively. 
The cells and nets of the circuit are represented as nodes $\mathcal{V}$, and multiple cells are connected via a net, forming an edge. 

{Traditionally, MOSFET-based netlists (e.g., Figure~\ref{fig:architecture}(a)) were represented as graphs by designing the three terminals (i.e., drain, gate, and source) as separate nodes~\cite{rubanov2006high} or by assigning 3-bit labels to the edges~\cite{kunal2020gana}.
To efficiently manage the number of nodes, we represent each MOSFET as a single node and assign distinct edge types to indicate connections to its three terminals.
Additionally, we use a directional graph to distinguish the directionality of influence, as the information exchanged between nodes within the circuit differs, especially at the supply voltage and ground.}
A comprehensive visualization of the graph representation of the circuit is presented in Figure~\ref{fig:Graph_Visulization}.
We provide detailed information about the node types and edge types as follows:

\noindent\textbf{- Node types.} Each node $v_i \in \mathcal{V}$ is associated with a feature vector $\mathbf{x}_i$, which is the $i$-th row of $\mathbf{X}$. We set $\mathbf{x}_i$ as a one-hot vector for each node type, and the assigned values for each node type are shown in the upper part of Table~\ref{tab:node/edge types}.

\noindent\textbf{- Edge types.} Each edge $e_{ij} \in \mathcal{E}$, which connects nodes $v_i$ and $v_j$, has an edge type $r$. Each edge type $r$ is designed to distinguish between the edge from the cell to the net and the edge from the net to the cell, which represents the directionality of the influence exchanged between nodes in a directed graph. For example, if a MOSFET is connected to a general net, both directional edges between the MOSFET and the net are created. However, if connected to a supply voltage (i.e., type $0$) or ground (i.e., type $1$), only the edge from the supply voltage or ground to the MOSFET is created to ensure that the node representations for ground or supply voltage are not affected by connected components. Specifically, we set edge types 0 to 13 to represent edges directed from the net to the cell, and types 14 to 27 to represent edges directed from the cell to the net. The former is based on the lower part of Table~\ref{tab:node/edge types}, and the latter follows the same order of node types as the former.

\noindent \textbf{Notations.} To detect the target circuit within the entire circuit, we convert both the target circuit and the entire circuit into a target graph and an entire graph, respectively. We define the target graph as $\mathcal{G}_T = (\mathcal{V}_T, \mathcal{E}_T, \mathbf{A}_T, \mathbf{X}_T )$ and the entire graph as $\mathcal{G}_E = (\mathcal{V}_E, \mathcal{E}_E, \mathbf{A}_E, \mathbf{X}_E )$, respectively.
{Figure~\ref{fig:architecture}(b) provides examples of the target graph and the entire graph.}

\subsection{$K$-hop Subgraph Formation}\label{subsec:K-hop_Subgraph_Formation}

Existing matching methods encounter time complexity issues when applied to large-scale circuits, as they optimize a matching matrix that includes node-to-node matching between two graphs. Specifically, the size of the matching matrix is $|\mathcal{V}_T|\times  |\mathcal{V}_E|$, which leads to significantly reduced time efficiency as $|\mathcal{V}_E|$ becomes larger in large-scale circuits. To address this issue, we propose a GNN-based method to predict regions with a high probability of containing the target graph (See details in Sec~\ref{subsec:Target_Region_Prediction}). In this section, we introduce how we extracted positive and negative samples from the entire graph to train the GNN.
The positive samples are $k$-hop subgraphs that contain $\mathcal{G}_T$, while the negative samples are subgraphs that do not contain $\mathcal{G}_T$. 

\smallskip
\noindent\textbf{Positive samples.} To ensure the diversity of positive samples and to minimize the size difference between the positive samples and  $\mathcal{G}_T$, we set the value of $K$ to the radius of $\mathcal{G}_T$ (i.e., half of the diameter). This is because the radius of $\mathcal{G}_T$ represents the minimum $K$-value that allows the $K$-hop subgraph to include $\mathcal{G}_T$. We extract various $K$-hop subgraphs that fully contain $\mathcal{G}_T$ from the entire graph as positive samples. In other words, the formation of a positive $K$-hop subgraph $\mathcal{G}_P$ from $\mathcal{G}$ can be defined as follows:
\begin{equation}\label{eqn:Gp}
\small
\mathcal{G}_P=\text{Subgraph}_K(\mathcal{G},v) \enspace \text{such that} \enspace K=\text{Radius}(\mathcal{G}_T) \; \text{and} \; \mathcal{G}_T \subseteq \mathcal{G}_P,
\end{equation}
where $\text{Subgraph}_K(\mathcal{G},v)$ represents the $K$-hop neighborhood subgraph of $\mathcal{G}$ centered at a node $v$ that includes all nodes and edges within $K$ hops of $v$.

\smallskip
\noindent\textbf{Negative samples.}
To effectively train the GNN to predict whether $k$-hop graphs fully contain $\mathcal{G}_T$, we define four types of negative samples $\mathcal{G}_N$ as follows:
\vspace{2mm}
\begin{enumerate}[label=\arabic*)]
\item \textbf{Partial} : $K$-hop subgraphs that partially contain the target graph.
\item \textbf{Mutation} : $K$-hop subgraphs generated from positive samples by converting P-mos to N-mos and N-mos to P-mos, or by swapping the source and drain terminals.
\item \textbf{Others} : $K$-hop subgraphs that contain different target circuits.
\item \textbf{Random} : Randomly sampled $K$-hop subgraphs, except for the positive samples. 
\end{enumerate}
It is important to note that we use hard negative samples, such as Partial and Mutation, to filter out $K$-hop subgraphs that partially contain or are slightly modified from the target circuit. This allows us to focus on predicting $K$-hop subgraphs that fully contain the target circuit.

\smallskip
\noindent{\textbf{$K$-hop subgraph formation for final matching.}
After completing all training, for the final target circuit matching, the $K$-hop subgraph formation is performed by considering each node in the entire graph as a center node.
Figure~\ref{fig:architecture}(c) illustrates the process of $K$-hop subgraph formation for final target circuit matching.}

\subsection{Target Region Prediction}\label{subsec:Target_Region_Prediction}
We train a GNN-based classifier to predict the presence of the target graph within $K$-hop subgraphs. 
As shown in Section~\ref{subsec:ablation_studies}, we observed that applying a GNN separately to each K-hop graph to generate individual embeddings results in significant time consumption.
To address this issue, we propose to apply the GNN \textit{only once} to the entire graph and derive the embeddings of various K-hop graphs from the precomputed embeddings of the entire graph, instead of applying the GNN separately to each K-hop subgraph.
 The main idea is to sample from the node embeddings of the entire graph, which will be explained in detail later in Sec~\ref{2.3.2}.
In summary, we first obtain node embeddings from the entire graph (Sec~\ref{2.3.1}), extract $K$-hop subgraph embeddings from the node embeddings of the entire graph (Sec~\ref{2.3.2}), and finally predict whether each extracted $K$-hop subgraph contains the target graph based on the embedding of the $K$-hop subgraph (Sec~\ref{2.3.3}).

\vspace{2mm}
\subsubsection{Obtaining Node Embeddings from Entire Graph}
\label{2.3.1}
Using the node features and edge types defined in Sec~\ref{subsec:Netlist to graph}, we establish the node embedding updates for the target graph $\mathcal{G}_T$ and the entire graph $\mathcal{G}_E$ within the framework of \cite{schlichtkrull2018modeling}.
The initial node embedding for each node $v_i$ is defined using its node feature $\mathbf{x}_i$. The embedding update in the first layer is as follows:
\vspace{-2mm}
\begin{equation}\label{eqn:hi0}
h_{i}^{(0)} = \mathbf{x}_{i}.
\end{equation}
In subsequent layers, this initial embedding $h_i^{(0)}$ is iteratively updated. To compute the forward pass update for each node $v_i$ in a graph with multiple edge types, we define the propagation model as follows:
\begin{equation}\label{eqn:hi(l+1)}
h_{i}^{(l+1)} = \sigma \left( \sum_{r=0}^{27} \sum_{j \in \mathcal{N}^r_{i}} \frac{1}{c_{i, r}} W_r^{(l)} h_{j}^{(l)} + W_0^{(l)} h_{i}^{(l)} \right),
\end{equation}
where $h_i^{(l)}$ represents the hidden representation of node $v_i$ in the $l$-th layer. $\mathcal{N}^r_{i}$ denotes the index set of neighboring nodes of node $i$ based on the edge type $r \in R$ and $c_{i,r}$ is a normalization constant specific to the problem such as $c_{i,r}=|\mathcal{N}^r_{i}|$. 


The main challenge in applying Equation~\ref{eqn:hi(l+1)} to multi-relational data is that the number of parameters increases rapidly as the number of edge types in the graph grows, which can lead to overfitting on infrequent relations and a substantial increase in model size. To address this, we employ basis decomposition as a method for regularizing the layer weights as follows:
\begin{equation}
\label{eq:basis}
\vspace{-1mm}
W_r^{(l)} = \sum_{b=1}^B a_{rb}^{(l)} V_b^{(l)},
\end{equation}
where $V_b^{(l)}$ is a basis transformation matrix shared across all edge types, and $a_{rb}^{(l)}$ is a coefficient defined for each edge type $r$ and basis matrix $V_b^{(l)}$ that adjusts the influence of the basis matrix for each edge type $r$.

\vspace{3mm}

\subsubsection{Extracting K-hop Subgraph Embeddings and Target Graph Embeddings}
\label{2.3.2}
\begin{figure}[t] 
\begin{center}
\includegraphics[width=1\linewidth]{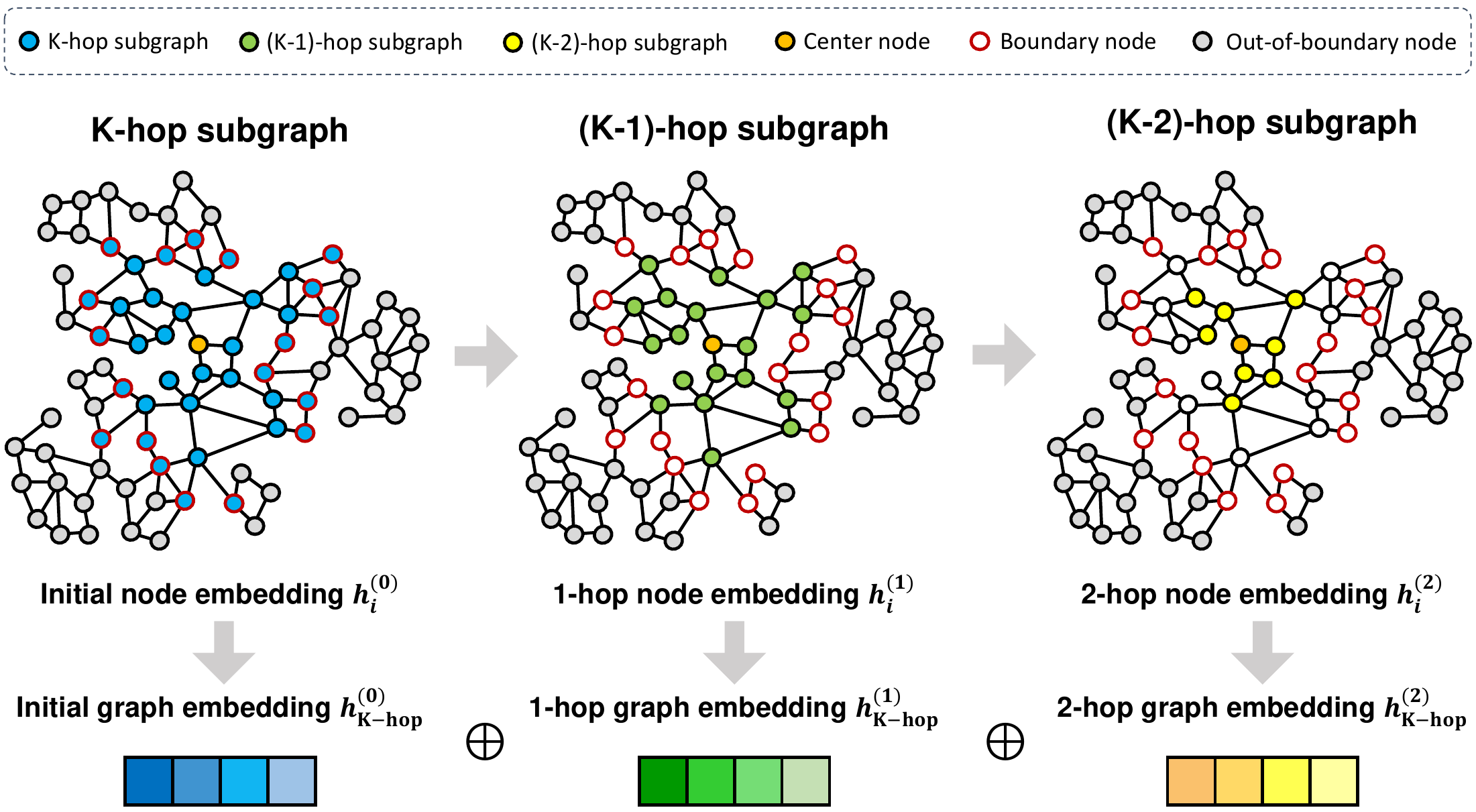}
\end{center}
\vspace{-1mm}
\caption{
Extracting K-hop subgraph embeddings from entire graph}
\label{fig:subgraph_embedding}
\vspace{-2mm}
\end{figure}

To derive the $K$-hop subgraph embedding from the node embeddings of the entire graph, it is essential to exclude information from nodes outside the boundary of the $K$-hop subgraph.
For this purpose, for each node $v_i$, we utilize all embeddings, including the initial node embedding $h_{i}^{(0)}$, the 1-hop node embedding $h_{i}^{(1)}$, and up to the L-hop node embedding $h_{i}^{(L)}$.
{Figure~\ref{fig:subgraph_embedding} illustrates the process for obtaining $K$-hop subgraph embeddings.}

We first sample a $(K-l)$-hop subgraph from the center node $v_c$ of the $K$-hop subgraph in the entire graph $\mathcal{G}_E$ as:
\begin{equation}
\mathcal{V}_{K}^{l}(v_c) = \{ v \in \mathcal{V}_E \mid \text{distance}(v_c, v) \leq K-l \},
\end{equation}
where $\mathcal{V}_{K}^{l}(v_c)$ denotes the set of all nodes included in the $(K-l)$ hop subgraph centered around the center node $v_c$. Specifically, $v_c$ represents the pre-defined center node for each positive and negative sample during the training phase and corresponds to all nodes within the entire graph during the testing phase.
Subsequently, we obtain the $l$-hop graph embedding $h_{\text{$K$-hop}}^{(l)}$ of the $K$-hop subgraph by pooling the $l$-hop embeddings $ h^{(l)}_i$ of nodes belonging to $\mathcal{V}_{K}^{l}(v_c)$ as:

\begin{equation}
h_{\text{$K$-hop}}^{(l)} = \text{Pool} \left( \{ h^{(l)}_i \mid v_i \in \mathcal{V}_{K}^{l}(v_c) \} \right),
\end{equation}
 where $\text{Pool}(\cdot)$ is the sum pooling function that aggregates the individual node embeddings. 
 It is important to note that $h_{\text{$K$-hop}}^{(l)}$ is designed to exclude information beyond the boundary of the $K$-hop subgraph (i.e., the gray nodes in Figure~\ref{fig:subgraph_embedding}).
This is because the information from nodes outside the boundary of the $K$-hop subgraph exists only in $\{ h^{(l)}_i \mid v_i \in \mathcal{V}_{E} \setminus \mathcal{V}_{K}^{l}(v_c) \}$.
 
 To obtain the final $K$-hop subgraph embedding, we concatenate $h_{\text{$K$-hop}}^{(l)}$ from all hops, ranging from $l=0$ to $l=L$ as:
\vspace{-1mm}
\begin{equation}
h_{\text{$K$-hop}} = \text{Concat} \left( h_{\text{K-hop}}^{(0)}, h_{\text{$K$-hop}}^{(1)}, \dots, h_{\text{$K$-hop}}^{(L)} \right),
\end{equation}
where $h_{\text{$K$-hop}}$ represents the final $K$-hop subgraph embedding, which, like each $h_{\text{$K$-hop}}^{(l)}$, does not include any information beyond the boundary nodes.

Similarly, we construct the $l$-hop graph embedding $h_{\text{target}}^{(l)}$ of the target graph by pooling the $l$-hop embeddings $ h^{(l)}_i$ as:

\begin{equation}
h_{\text{target}}^{(l)} = \text{Pool} \left( \{ h^{(l)}_i \mid v_i \in \mathcal{V}_{T} \} \right).
\end{equation}
Finally, we obtain the final target graph embedding by concatenating $h_{\text{target}}^{(l)}$ from all hops, ranging from $l=0$ to $L$ as:

\begin{equation}
h_{\text{target}} = \text{Concat} \left( h_{\text{target}}^{(0)}, h_{\text{target}}^{(1)}, \dots, h_{\text{target}}^{(L)} \right).
\end{equation}

\subsubsection{Predicting the Presence of $\mathcal{G}_T$ within $K$-hop Subgraphs}
\label{2.3.3}

We predict the presence of the target graph within the $K$-hop subgraph based on the representations obtained in Sec~\ref{2.3.2}.
We concatenate the $K$-hop subgraph representation $h_{K\text{-hop}}$ with the target graph representation $h_{\text{target}}$ and use an MLP for the final prediction as follows:
\begin{equation}
\label{eq:predicted_probability}
 \hat{p}=\text{MLP}\left ( h_{\text{concat}} \right ) = \text{MLP}\left ( \left [ h_{K\text{-hop}};h_{\text{target}} \right ] \right ),
\end{equation}
where $\hat{p} \in [0, 1]$ represents the predicted probability that the target graph is present within the $K$-hop subgraph.
We optimize the binary cross-entropy loss to train the model as:
\begin{equation}
\label{eq:loss}
\mathcal{L}(y, \hat{p}) = - \frac{1}{N} \sum_{i=1}^{N} \left[ y_i \log(\hat{p}_i) + (1 - y_i) \log(1 - \hat{p}_i) \right],
\end{equation}
where $N$ represents the total number of positive and negative samples used for training. The ground truth label $y_i$ is assigned as $1$ for positive samples $\mathcal{G}_P$ where the target graph $\mathcal{G}_T$ is present, and $0$ for negative samples $\mathcal{G}_N$ where it is absent. {Figure~\ref{fig:architecture}(d) illustrates the process of target region prediction.}

\begin{table*}[ht]
\centering
\renewcommand{\arraystretch}{1.1}
\resizebox{0.7\textwidth}{!}{%
\begin{tabular}{cc|cc|cc}
\hline
                                 &                                  & \multicolumn{2}{c|}{VF2}            & \multicolumn{2}{c}{Ours}                                              \\
\multirow{-2}{*}{Entire Circuit} & \multirow{-2}{*}{Target Circuit} & \# of Matching & Time (seconds)              & \# of Matching             & Time (seconds)                \\ \hline
                                 & \textbf{\textit{nand5}}                            & 576            & 1330.32 \scriptsize{\textbf{(78.06\%)}}                                 & 576 & \textbf{291.84} \\
                                 & \textbf{\textit{nand3}}                            & 4              & 546.69 \scriptsize{\textbf{(82.35\%)}}  & 4   & \textbf{96.46}  \\
                                 & \textbf{\textit{col\_sel1m}}                       & 512            & 1585.33 \scriptsize{\textbf{(92.49\%)}} & 512 & \textbf{118.93} \\
\multirow{-4}{*}{\textbf{\textit{col\_sel9\_512}}} & \textbf{\textit{decoder2\_4}}                      & 1              & 301.15 \scriptsize{\textbf{(15.45\%)}}  & 1   & \textbf{254.62} \\ \hline
                                 & \textbf{\textit{nand5}}                            & 576            & 1312.86 \scriptsize{\textbf{(76.80\%)}} & 576                        & \textbf{304.53}                        \\
                                 & \textbf{\textit{nand3}}                            & 4              & 526.08 \scriptsize{\textbf{(82.44\%)}}  & 4                          & \textbf{92.34}                         \\
                                 & \textbf{\textit{col\_sel1m}}                       & 512            & 1481.52 \scriptsize{\textbf{(91.93\%)}} & 512                        & \textbf{119.45}                        \\
                                 & \textbf{\textit{decoder2\_4}}                      & 1              & 332.37 \scriptsize{\textbf{(14.48\%)}}  & 1                          & \textbf{284.21}                        \\
\multirow{-5}{*}{\textbf{\textit{deco9\_512}}}     & \textbf{\textit{deco1m}}                           & 512            & 1555.30 \scriptsize{\textbf{(91.64\%)}} & 512                        & \textbf{129.88}                        \\ \hline
                                 & \textbf{\textit{ctg}}                              & 2              & 113.34 \scriptsize{\textbf{(8.51\%)}}   & 2                          & \textbf{103.69}                        \\
                                 & \textbf{\textit{inv}}                              & 2616           & 238.02 \scriptsize{\textbf{(2.81\%)}}   & 2616                       & \textbf{231.33}                        \\
                                 & \textbf{\textit{nand2}}                            & 6              & 858.56 \scriptsize{\textbf{(69.69\%)}}  & 6                          & \textbf{260.21}                        \\
                                 & \textbf{\textit{nor3}}                             & 1              & 683.54 \scriptsize{\textbf{(20.26\%)}}   & 1                          & \textbf{545.03}                        \\
                                 & \textbf{\textit{nand5}}                            & 1153           & 6169.48 \scriptsize{\textbf{(87.79\%)}} & 1153                       & \textbf{753.23}                        \\ 
                                 & \textbf{\textit{col\_sel1m}}                       & 1024           & 6721.20 \scriptsize{\textbf{(96.47\%)}} & 1024                       & \textbf{237.23}                         \\
                                 & \textbf{\textit{nand3}}                            & 9              & 2502.03 \scriptsize{\textbf{(92.11\%)}} & 9                          & \textbf{197.30}                        \\
                                 & \textbf{\textit{decoder2\_4}}                      & 2              & 1336.40 \scriptsize{\textbf{(63.42\%)}} & 2                          & \textbf{488.82}                        \\
\multirow{-9}{*}{\textbf{\textit{ctrl256kbm}}}     & \textbf{\textit{deco1m}}                           & 1024           & 6773.05 \scriptsize{\textbf{(96.38\%)}} & 1024                       & \textbf{244.70}                        \\ \hline
\end{tabular}
}
\vspace{-1mm}
\caption{Running time (sec) for matching all target circuits. 
We report the reduction rate of our method compared to VF2 in parentheses.}
\label{table:matching_all_target}
\vspace{-2mm}
\end{table*}

\subsection{Target Graph Matching} \label{subsec:Matching}
Recall that our goal is to identify the locations of the target graphs $\mathcal{G}_{T}$ within the entire graph $\mathcal{G}_{E}$. 
We generate $K$-hop subgraph embeddings $h_{\text{$K$-hop}}$ centered around all nodes in the entire graph $\mathcal{G}_{E}$ and compute the predicted probability $\hat{p}$ for each $K$-hop subgraph based on $h_{\text{$K$-hop}}$. 
We arrange the computed predicted probabilities $\hat{p}$ in descending order, prioritizing those subgraphs for which the model predicts a higher likelihood of containing the target graph $\mathcal{G}_{T}$. 
Subsequently, we apply the VF2 algorithm~\cite{cordella2004sub} to match the target graph within each K-hop subgraph based on the sorted order. Figure ~\ref{fig:architecture}(d) illustrates the results of target graph matching.

\section{Experiments}
\noindent\textbf{Datasets. }
We use the \textbf{\textit{Lassen DRAMB}} dataset, which includes various circuits comprising the \textbf{\textit{Dram1gbn}} circuit, for this experiment.
The dataset consists of MOSFET-based netlists, with each netlist containing information about the cells that compose the circuits and the nets that connect these cells.
The cells in this dataset include P-MOS, N-MOS, capacitors, resistors, and inductors.
The number of nodes for each circuit converted into a graph is presented in Table~\ref{tab:datastat}. 

\begin{table}[H]
\centering
\resizebox{0.75\columnwidth}{!}{%
\begin{tabular}{cccc}
\toprule
Circuit        & \# of Nodes & Circuit        & \# of Nodes \\ \midrule
\textbf{\textit{ctrl256kbm}}     & 26430       & \textbf{\textit{deco9\_512}}  & 12343       \\
\textbf{\textit{col\_sel9\_512}} & 12343       & \textbf{\textit{deco7\_128}}  & 3077        \\
\textbf{\textit{col\_sel7\_128}} & 3077        & \textbf{\textit{atd18m}}      & 1529        \\
\textbf{\textit{atd18}}          & 1525        & \textbf{\textit{array512x1}}  & 520         \\
\textbf{\textit{decoder4\_16}}   & 326         & \textbf{\textit{deco3\_8m}}   & 181         \\
\textbf{\textit{col\_sel3\_8m}}  & 181         & \textbf{\textit{xvald}}       & 100         \\
\textbf{\textit{atd1}}           & 79          & \textbf{\textit{decoder2\_4}} & 72          \\
\textbf{\textit{io}}             & 62          & \textbf{\textit{sen\_ena}}    & 40          \\
\textbf{\textit{delay10}}        & 34          & \textbf{\textit{deco1m}}      & 30          \\
\textbf{\textit{col\_sel1m}}     & 30          & \textbf{\textit{rl\_sel}}     & 26          \\
\textbf{\textit{nand5}}          & 23          & \textbf{\textit{delay6}}      & 22          \\
\textbf{\textit{main\_sense}}    & 20          & \textbf{\textit{nor4}}        & 19          \\
\textbf{\textit{delay4}}         & 16          & \textbf{\textit{nand3}}       & 15          \\
\textbf{\textit{nor3}}           & 15          & \textbf{\textit{nand2}}       & 11          \\
\textbf{\textit{precharge}}      & 8           & \textbf{\textit{inv}}         & 7           \\
\textbf{\textit{ctg}}            & 8           &                               &             \\ \bottomrule
\end{tabular}
}
\vspace{-1mm}
\caption{Statistics of \textbf{\textit{Lassen DRAMB}} used for experiments.}
\label{tab:datastat}
\vspace{-1mm}
\end{table}

\noindent\textbf{Evaluation Protocol. }
 We set the number of layers $L$ to 2, and the model is trained for $1000$ epochs using the Adam SGD optimizer with a learning rate of 0.001.
During training, we use 8 circuits, including \textbf{\textit{ctg, inv, delay10, nand2, nor3, nand5, nor4,}} and \textbf{\textit{nand3}}, as target circuits {as they the circuits that are evenly distributed across all circuits.}
Additionally, we utilize a total of 18 circuits, including \textbf{\textit{io, dff1,  main\_sense, atd1, decoder4\_16, decoder2\_4, deco3\_8m, delay4, rl\_sel, delay6, sen\_ena, delay16, xvald, atd18m, atd18, col\_sel1m, deco1m}} and \textbf{\textit{decoder6\_64}}, as entire circuits for training.
To evaluate matching performance, we use the three largest circuits of the dataset, including \textbf{\textit{ctrl256kbm, deco9\_512}} and \textbf{\textit{col\_sel9\_512}}, as the entire circuits, all of which are not seen during training.
The model is trained using positive samples and four types of negative samples 
derived from the 8 target circuits and 18 entire circuits.
During training, we use 200 positive samples, 100 partial samples, 50 mutation samples, 50 other samples, and 300 random samples for each target circuit. For the evaluation of target region prediction,  we use 100 positive samples and 100 random samples, which are considered as negatives.

\begin{table*}[ht]
\centering
\renewcommand{\arraystretch}{1.1}
\resizebox{0.75\textwidth}{!}{%
\begin{tabular}{cc|>{\centering\arraybackslash}p{3cm}>{\centering\arraybackslash}p{3cm}>{\centering\arraybackslash}p{2cm}}
\hline
Entire circuit               & Target Circuit & FastPFP   & NeuroMatch & Ours    \\ \hline
\multirow{4}{*}{\textbf{\textit{col\_sel9\_512}}} & \textbf{\textit{nand5}}          & 78718.52 \scriptsize{\textbf{(99.95\%)}}  & 13807.22 \scriptsize{\textbf{(99.71\%)}}   & \textbf{40.08}   \\
                                & \textbf{\textit{nand3}}          & 85888.94 \scriptsize{\textbf{(99.89\%)}}  & 7491.89 \scriptsize{\textbf{(98.76\%)}}    & \textbf{92.89}   \\
                                & \textbf{\textit{col\_sel1m}}     & 104901.07 \scriptsize{\textbf{(99.95\%)}} & 15995.53
 \scriptsize{\textbf{(99.73\%)}}   & \textbf{42.99}   \\
                                & \textbf{\textit{decoder2\_4}}    & 109795.59 \scriptsize{\textbf{(99.76\%)}} &   45601.25
 \scriptsize{\textbf{(99.44\%)}}                    & \textbf{254.62}  \\ \hline
\multirow{5}{*}{\textbf{\textit{deco9\_512}}}     & \textbf{\textit{nand5}}          & 74326.43 \scriptsize{\textbf{(99.94\%)}}  & 17243.17
 \scriptsize{\textbf{(99.76\%)}}   & \textbf{41.05}   \\
                                & \textbf{\textit{nand3}}          & 82536.47 \scriptsize{\textbf{(99.88\%)}}  & 12979.29
 \scriptsize{\textbf{(99.28\%)}}    & \textbf{92.29}   \\
                                & \textbf{\textit{col\_sel1m}}     & 110794.83 \scriptsize{\textbf{(99.96\%)}} & 19532.43
 \scriptsize{\textbf{(99.80\%)}}   & \textbf{38.79}   \\
                                & \textbf{\textit{decoder2\_4}}    & 104759.36 \scriptsize{\textbf{(99.72\%)}} &  99169.86 \scriptsize{\textbf{(99.71\%)}}                   & \textbf{284.21}  \\
                                & \textbf{\textit{deco1m}}         & 79264.99 \scriptsize{\textbf{(99.96\%)}}  &  18196.41 \scriptsize{\textbf{(99.83\%)}}                   & \textbf{29.27}   \\ \hline
\multirow{9}{*}{\textbf{\textit{ctrl256kbm}}}     & \textbf{\textit{ctg}}            & 74767.46 \scriptsize{\textbf{(99.91\%)}}  &    3915.78
 \scriptsize{\textbf{(98.36\%)}}                 & \textbf{63.97}   \\
                                & \textbf{\textit{inv}}            & 73295.37 \scriptsize{\textbf{(99.91\%)}}  &    3178.27 \scriptsize{\textbf{(98.08\%)}}                  & \textbf{60.85}   \\
                                & \textbf{\textit{nand2}}          & 150285.43 \scriptsize{\textbf{(99.94\%)}} &  6331.35 \scriptsize{\textbf{(98.68\%)}}                   & \textbf{83.28}   \\
                                & \textbf{\textit{nor3}}           & 73952.64 \scriptsize{\textbf{(99.26\%)}}  &   11071.35 \scriptsize{\textbf{(95.07\%)}}                  & \textbf{545.03}  \\
                                & \textbf{\textit{nand5}}          & 783472.46 \scriptsize{\textbf{(99.99\%)}} &           18744.59 \scriptsize{\textbf{(99.61\%)}}          & \textbf{71.90}   \\
                                & \textbf{\textit{col\_sel1m}}     & 792575.38 \scriptsize{\textbf{(99.98\%)}} &           20527.66 \scriptsize{\textbf{(99.32\%)}}          &  \textbf{139.04}                 \\
                                & \textbf{\textit{nand3}}          & 103758.36 \scriptsize{\textbf{(99.81\%)}} &           11515.49 \scriptsize{\textbf{(98.29\%)}}          & \textbf{195.77}  \\
                                & \textbf{\textit{decoder2\_4}}    & 82493.61 \scriptsize{\textbf{(99.90\%)}}  &    82496.3 \scriptsize{\textbf{(99.91\%)}}                 & \textbf{77.66}  \\
                                & \textbf{\textit{deco1m}}         & 819428.42 \scriptsize{\textbf{(99.98\%)}} &           20465.20 \scriptsize{\textbf{(99.32\%)}}          &    \textbf{137.61}                   \\ \hline
\end{tabular}
}
\vspace{-1mm}
\caption{Running time (sec) for matching one target circuit, with the reduction rate of our method relative to the baselines in parentheses.}
\label{table:matching_one_target}
\vspace{-4mm}
\end{table*}

\subsection{Matching Performance}  
\label{subsec:matching_performance}
\noindent\textbf{Overview.} In this section, we evaluate the matching performance on the three largest circuits in the dataset. Since our proposed method is based on exact matching, this experiment considers a matching to be successful only when all nodes and edges of the target circuit are matched exactly. Therefore, the matching performance in this experiment is measured based on the time required to achieve a successful exact matching.
First, we compare the time required to match \textit{all target circuits} contained in the entire circuits with the VF2~\cite{cordella2004sub}, which, as noted in ~\cite{liu2020neural}, does not involve heuristic rules (e.g., rules in \cite{carletti2017challenging}).
Additionally, we compare the time required to match \textit{one target circuit} within the entire circuits against baseline methods designed for single target circuit matching.
We observed that applying ISONET~\cite{roy2022interpretable} to large-scale circuits results in out-of-memory issues with inefficient execution times.
This occurs because the size of the edge permutation matrix (i.e., matching matrix) introduced for matching in [10] increases with the number of graph nodes, leading to a rapid increase in memory usage. 
This clearly demonstrates the inefficiency of the matching matrix in large-scale circuits, as mentioned in Section~\ref{subsec:K-hop_Subgraph_Formation}.
Therefore, we adopt FastPFP~\cite{lu2016fast} and NeuroMatch~\cite{lou2020neural} as baselines for this task.

\noindent\textbf{Baselines.}
In this section, we provide details of the baseline methods used in the experiments.
\begin{itemize}[leftmargin=3mm]

    \item \textbf{VF2} \cite{cordella2004sub} proposes a deterministic matching method aimed at identifying both isomorphism and subgraph isomorphism. It introduces a State Space Representation (SSR) in the matching process and incorporates various rules to efficiently explore the search tree. Additionally, it optimizes the organization of data structures used during the search process to improve both time and space efficiency.
    \item \textbf{FastPFP} \cite{lu2016fast} proposes a new projected fixed-point method for large-scale graph matching. It addresses the issue of high computational complexity by using doubly stochastic projection and demonstrates scalability for large graphs with more than 1,000 nodes.
    \item \textbf{NeuroMatch} \cite{lou2020neural} trains a GNN to capture geometric constraints related to subgraph relationships and performs subgraph matching in the embedding space. It uses order embeddings to represent each node in the embedding space, ensuring that the hierarchical relationships of subgraphs are accurately reflected in the embedding space.

\end{itemize}

\noindent \textbf{Experiment Results.}
Table~\ref{table:matching_all_target} shows the experimental results for running time for matching all target circuits.
We have the following observations: 
\textbf{1)} Our method produces the same number of matchings as VF2 for all target circuits, indicating that it successfully matches all target circuits.
\textbf{2)} Our method achieves a reduced running time compared to VF2, demonstrating that it provides higher time efficiency for matching. Consequently, our method ensures both accuracy and high time efficiency by successfully matching all target circuits and significantly reducing running time. 
\textbf{3)} We provide the reduction rate of our method relative to VF2 next to the VF2 running time. 
Our approach achieves up to a 96.47\% reduction in running time compared to VF2, indicating that it can achieve significantly more efficient matching in large-scale circuits.

In Table~\ref{table:matching_one_target}, we have also presented the experimental results for running time for matching one target circuits.
We have the following observations: \textbf{1)} Our method achieves the shortest time for identifying a single circuit across all target circuits when compared to FastPFP.
\textbf{2)} FastPFP shows highly inefficient running time. This inefficiency arises from FastPFP's progressive projection method, which relies on matrix computations and requires performing complex operations at each iteration. Notably, the partial doubly stochastic matrix projection significantly increases computational complexity as the matrix size grows. Consequently, this leads to a significant increase in matching time for large-scale circuits.
\textbf{3)} NeuroMatch shows relatively greater efficiency compared to FastPFP, but it still encounters inefficiencies when applied to large-scale circuits. 
This is because the node embeddings learned by NeuroMatch fail to effectively capture subgraph relationships, and the voting algorithm used for matching is inefficient for large-scale circuits.
\textbf{4)} Our method achieves reduction rates of over 98\% in most cases across all target circuits compared to the baselines. This demonstrates that, in contrast to the inefficiency of FastPFP and NeuroMatch for large-scale circuits, our approach performs highly efficient matching, making it highly suitable for practical applications.


\begin{table*}[t]
\renewcommand{\arraystretch}{1.1}
\centering
\resizebox{\textwidth}{!}{%
\begin{tabular}{>{\centering\arraybackslash}p{1.4cm}
                >{\centering\arraybackslash}p{2.0cm}
                >{\centering\arraybackslash}p{1.3cm}
                >{\centering\arraybackslash}p{1.3cm}
                >{\centering\arraybackslash}p{1.3cm}
                >{\centering\arraybackslash}p{1.3cm}
                >{\centering\arraybackslash}p{1.3cm}
                >{\centering\arraybackslash}p{1.3cm}
                >{\centering\arraybackslash}p{1.3cm}
                >{\centering\arraybackslash}p{1.3cm}
                >{\centering\arraybackslash}p{1.3cm}} 
\hline
                            &      Target Circuit    & \textbf{\textit{ctg}}                 & \textbf{\textit{inv}}                 & \textbf{\textit{delay10}}               & \textbf{\textit{nand2}}                & \textbf{\textit{nor3}}                  & \textbf{\textit{nand5}}                & \textbf{\textit{nor4}}                  & \textbf{\textit{nand3}}                 & ALL                 \\ \hline
\multirow{2}{*}{GCN}        & Accuracy & 72.99\scriptsize{±0.00}   & 54.75\scriptsize{±3.25}   & 61.04\scriptsize{±5.80}    & 89.38\scriptsize{±0.88}   & 91.36\scriptsize{±0.45}    & 88.00\scriptsize{±1.50}   & 79.94\scriptsize{±1.97}    & 75.94\scriptsize{±1.97}    & 75.99\scriptsize{±1.94}   \\
                            & AUROC    & 59.13\tiny{±1.92}   & 75.39\scriptsize{±0.43}   & 70.73\scriptsize{±5.00}    & 93.03\scriptsize{±6.65}   & 97.40\scriptsize{±2.20}    & 99.82\scriptsize{±0.08}   & 99.14\scriptsize{±0.85}    & 97.19\scriptsize{±0.62}    & 92.23\scriptsize{±0.02}   \\
\multirow{2}{*}{GIN}        & Accuracy & 72.99\scriptsize{±0.00}   & 55.75\scriptsize{±1.75}   & 58.56\scriptsize{±2.76}    & 88.93\scriptsize{±0.44}   & 90.91\scriptsize{±0.00}    & 94.25\scriptsize{±2.25}   & 76.76\scriptsize{±1.41}    & 88.41\scriptsize{±4.80}    & 77.30\scriptsize{±1.35}   \\
                            & AUROC    & 65.65\tiny{±3.89}   & 75.41\scriptsize{±3.27}   & 73.58\scriptsize{±5.71}    & 98.07\scriptsize{±1.84}   & 98.35\scriptsize{±1.05}    & 99.45\scriptsize{±0.54}   & 99.87\scriptsize{±0.13}    & 99.53\scriptsize{±0.32}    & 93.30\scriptsize{±0.45}   \\
\multirow{2}{*}{GAT}        & Accuracy & 72.99\scriptsize{±0.00}   & 52.83\scriptsize{±2.01}   & 55.24\scriptsize{±0.00}    & 88.49\scriptsize{±0.00}   & 90.91\scriptsize{±0.00}    & 79.57\scriptsize{±2.63}   & 79.57\scriptsize{±2.63}    & 78.34\scriptsize{±8.37}    & 75.07\scriptsize{±1.10}   \\
                            & AUROC    & 55.33\scriptsize{±3.65}   & 74.88\scriptsize{±3.08}   & 74.50\scriptsize{±4.10}    & 99.12\scriptsize{±0.56} & 96.93\scriptsize{±1.83} & 99.78\scriptsize{±0.12}   & 99.57\scriptsize{±0.43}    & 95.50\scriptsize{±5.50}    & 92.09\scriptsize{±1.29}   \\
\multirow{2}{*}{SimGNN}        & Accuracy & 48.90\scriptsize{±3.78}    & 53.50\scriptsize{±7.46}    & 43.64\scriptsize{±2.53}    & 53.09\scriptsize{±3.85}    & 59.09\scriptsize{±0.00}     & \textbf{99.49\scriptsize{±0.72}}    & 82.11\scriptsize{±4.38}     & 92.84\scriptsize{±0.65}    & 68.69\scriptsize{±6.43}   \\
                            & AUROC    & 50.42\scriptsize{±7.67}    & 55.21\scriptsize{±15.33}    & 40.22\scriptsize{±3.44}    & 88.41\scriptsize{±9.32}  & 86.56\scriptsize{±5.12}  & \textbf{100.00\scriptsize{±0.00}}    & 93.58\scriptsize{±2.10}     & 99.97\scriptsize{±0.04}   & 74.08\scriptsize{±12.81}  \\
\multirow{2}{*}{GMN-\footnotesize{embed}}       & Accuracy & 50.85\scriptsize{±2.22}   & 39.50\scriptsize{±5.41}   & 55.80\scriptsize{±0.00}    & 42.77\scriptsize{±1.02}   & 41.51\scriptsize{±1.05}    & 8.66\scriptsize{±2.88}   & 21.83\scriptsize{±2.44}    & 91.71\scriptsize{±2.61}    & 37.51\scriptsize{±3.25}   \\
                            & AUROC    & 51.61\scriptsize{±0.79}   & 38.87\scriptsize{±6.73}   & 60.76\scriptsize{±0.58}    & 26.10\scriptsize{±2.92} & 12.33\scriptsize{±6.12} & 4.08\scriptsize{±2.32}   & 7.31\scriptsize{±4.97}    & 96.38\scriptsize{±1.06}    & 34.94\scriptsize{±3.16}   \\
\multirow{2}{*}{GMN-\footnotesize{match}}        & Accuracy & 51.82\scriptsize{±5.25}   & 42.33\scriptsize{±7.51}   & 56.72\scriptsize{±4.63}    & 51.91\scriptsize{±6.40}   & 58.48\scriptsize{±0.52}    & 88.40\scriptsize{±1.28}   & 77.46\scriptsize{±1.21}    & 92.65\scriptsize{±0.56}    & 72.26\scriptsize{±4.11}   \\
                            & AUROC    & 48.95\scriptsize{±6.74}   & 38.51\scriptsize{±5.07}   & 58.28\scriptsize{±3.71}    & 93.64\scriptsize{±2.76} & 87.43\scriptsize{±6.33} & 95.09\scriptsize{±0.73}   & 92.97\scriptsize{±4.48}    & 99.94\scriptsize{±0.09}    & 79.67\scriptsize{±5.60}   \\
\multirow{2}{*}{NeuroMatch} & Accuracy & 61.55\scriptsize{±9.06}   & 57.66\scriptsize{±10.27}  & 49.72\scriptsize{±5.06}    & 89.08\scriptsize{±0.51}   & 91.82\scriptsize{±1.82}    & 94.97\scriptsize{±2.64}   & 90.14\scriptsize{±3.73}    & 95.85\scriptsize{±1.30}    & 77.79\scriptsize{±2.49}   \\
                            & AUROC    & 47.41\scriptsize{±5.64}   & 63.57\scriptsize{±4.26}  & 45.66\scriptsize{±5.87}    & 88.48\scriptsize{±8.58}   & 93.66\scriptsize{±3.01}    & 99.23\scriptsize{±0.66}   & 97.58\scriptsize{±1.55}    & 99.82\scriptsize{±0.30}    & 85.13\scriptsize{±1.26}   \\
\multirow{2}{*}{ISONET}      & Accuracy & 48.66\scriptsize{±2.75}   & 59.57\scriptsize{±2.76}   & 51.93\scriptsize{±4.97}    & 57.81\scriptsize{±4.17}   & 57.27\scriptsize{±1.82}    & 99.49\scriptsize{±0.51}   & 79.10\scriptsize{±0.41}    & 93.22\scriptsize{±0.00}    & 77.98\scriptsize{±1.17}   \\
                            & AUROC    & 48.85\tiny{±6.12}   & 60.19\scriptsize{±4.82}   & 55.38\scriptsize{±4.58}    & 88.56\scriptsize{±12.92}   & 91.53\scriptsize{±6.31}    & 99.99\scriptsize{±0.02}   & 98.15\scriptsize{±2.44}    & 99.66\scriptsize{±0.57}    & 89.85\scriptsize{±0.95}   \\
\multirow{2}{*}{Ours}       & Accuracy & \textbf{73.72\scriptsize{±0.00}} & \textbf{83.50\scriptsize{±1.00}} & \textbf{98.34\scriptsize{±0.55}} & \textbf{97.35\scriptsize{±0.83}} & \textbf{99.54\scriptsize{±0.45}} & 94.50\scriptsize{±0.50} & \textbf{90.49\scriptsize{±0.35}} & \textbf{100.00\scriptsize{±0.00}} & \textbf{91.98\scriptsize{±0.08}} \\
                            & AUROC    & \textbf{73.46\scriptsize{±1.31}} & \textbf{95.67\scriptsize{±0.44}} & \textbf{100.00\scriptsize{±0.00}} & \textbf{99.27\scriptsize{±0.73}} & \textbf{100.00\scriptsize{±0.00}} & \textbf{100.00\scriptsize{±0.00}} & \textbf{100.00\scriptsize{±0.00}} & \textbf{100.00\scriptsize{±0.00}} & \textbf{98.58\scriptsize{±0.04}} \\ \hline
\end{tabular}%
}
\vspace{-1mm}
\caption{Performance on the target region prediction task.}
\label{table:target_region_prediction}
\vspace{-2mm}
\end{table*}

\subsection{Target Region Prediction Performance}  \label{subsec:GNN_performance}
\noindent\textbf{Overview.} In this section, we conduct experiments on the target region prediction, which enables efficient matching by focusing on regions with a high likelihood of containing the target circuit.
For comparison, we use widely-used models such as GCN~\cite{kipf2016semi}, GIN~\cite{xu2018powerful}, and GAT~\cite{velivckovic2017graph}, as well as models designed for predicting relevance scores or subgraph isomorphism, such as SIMGNN~\cite{bai2019simgnn}, GMN~\cite{li2019graph}, NeuroMatch~\cite{lou2020neural}, and ISONET~\cite{roy2022interpretable}, as baselines. 

\noindent\textbf{Baselines.}
In this section, we provide details of the baseline methods used in the experiments, excluding widely-used GNN models such as GCN~\cite{kipf2016semi}, GIN~\cite{xu2018powerful}, and GAT~\cite{velivckovic2017graph}.
\begin{itemize}[leftmargin=3mm]

    
    
    \item \textbf{SimGNN} \cite{bai2019simgnn} formulates graph similarity computation as a learning task.
    Specifically, a new attention mechanism selects important nodes relevant to similarity to provide a global summary of the graph. To address the oversimplification of graph-level embeddings, pairwise node similarity scores are calculated to enhance the representation.

    \item \textbf{GMN} \cite{li2019graph} proposes a method for generating graph embeddings to efficiently reason about similarity. To compute similarity scores between graph pairs, GMN-$\text{\footnotesize{embed}}$ calculates embeddings independently for each graph in the pair, while GMN--$\text{\footnotesize{match}}$ computes embeddings based on a cross-attention layer applied to node pairs across the two graphs.

    \item \textbf{ISONET} \cite{roy2022interpretable} proposes an interpretable neural edge alignment formulation. It learns an optimal edge alignment plan after extracting edge embeddings from local neighborhoods. Edge-consistent mapping and asymmetric relevance scores enable accurate approximations of graph correspondences.
\end{itemize}


\noindent\textbf{Experiment Results.}
Table~\ref{table:target_region_prediction} shows the results for the target region prediction task.
We have the following observations: \textbf{1)} Our method achieves the highest accuracy across all target circuits, except for a single target circuit, \textbf{\textit{nand5}}. This indicates that our approach is the most effective in predicting target regions within each subgraph.
Additionally, we confirm that our $K$-hop subgraph embedding, constructed using all hops as described in Section~\ref{2.3.2}, achieves high prediction accuracy by effectively excluding information beyond the subgraph boundary.
\textbf{2)} Our method achieves the highest AUROC across all target circuits compared to other baselines. This demonstrates its effectiveness in performing final target circuit matching within the ranked regions.
\textbf{3)} Notably, our method achieves near-perfect accuracy and AUROC, approaching 100\%, for \textbf{\textit{nor3}} and \textbf{\textit{nand3}}. 
From this analysis, we observe a tendency for target graphs with a larger number of nodes to achieve higher prediction performance.
This is because smaller target graphs are more difficult to distinguish due to the presence of numerous similar surrounding regions. This finding suggests that simpler target graphs are more likely to have similar regions in large-scale circuits.


\subsection{Ablation Studies}  \label{subsec:ablation_studies}

\noindent\textbf{1) Our Subgraph Embedding Extraction} 

\begin{table}[t]
\centering
\begin{tabular}{cc|cc}
\hline
Entire circuit                  & Target Circuit & w/o our $h_{\mathcal{G}_{sub}}$ extraction & Ours              \\ \hline
\multirow{4}{*}{\textbf{\textit{col\_sel9\_512}}} & \textbf{\textit{nand5}}          & 480.38 \scriptsize{\textbf{(39.24\%)}}                       & \textbf{291.8421} \\
                                & \textbf{\textit{nand3}}          & 361.53 \scriptsize{\textbf{(73.32\%)}}                       & \textbf{96.4665}  \\
                                & \textbf{\textit{col\_sel1m}}     & 947.99  \scriptsize{\textbf{(87.45\%)}}                      & \textbf{118.9321} \\
                                & \textbf{\textit{decoder2\_4}}    & 947.86  \scriptsize{\textbf{(73.14\%)}}                      & \textbf{254.6227} \\ \hline
\multirow{5}{*}{\textbf{\textit{deco9\_512}}}     & \textbf{\textit{nand5}}          & 497.53 \scriptsize{\textbf{(38.79\%)}}                      & \textbf{304.5308} \\
                                & \textbf{\textit{nand3}}          & 384.46 \scriptsize{\textbf{(75.98\%)}}                      & \textbf{92.3401}  \\
                                & \textbf{\textit{col\_sel1m}}     & 982.38  \scriptsize{\textbf{(87.84\%)}}                     & \textbf{119.4577} \\
                                & \textbf{\textit{decoder2\_4}}    & 958.94 \scriptsize{\textbf{(70.36\%)}}                      & \textbf{284.2191} \\
                                & \textbf{\textit{deco1m}}         & 1036.94  \scriptsize{\textbf{(87.47\%)}}                    & \textbf{129.8804} \\ \hline
\multirow{9}{*}{\textbf{\textit{ctrl256kbm}}}     & \textbf{\textit{ctg}}            & 776.81 \scriptsize{\textbf{(86.65\%)}}                       & \textbf{103.6938} \\
                                & \textbf{\textit{inv}}            & 6042.31  \scriptsize{\textbf{(96.17\%)}}                     & \textbf{231.3311} \\
                                & \textbf{\textit{nand2}}          & 7871.31 \scriptsize{\textbf{(96.69\%)}}                     & \textbf{260.2116} \\
                                & \textbf{\textit{nor3}}           & 8443.96  \scriptsize{\textbf{(93.54\%)}}                     & \textbf{545.0347} \\
                                & \textbf{\textit{nand5}}          & 2113.13  \scriptsize{\textbf{(64.35\%)}}                    & \textbf{753.235}  \\
                                & \textbf{\textit{col\_sel1m}}     & 1862.69  \scriptsize{\textbf{(87.26\%)}}                    & \textbf{237.233}  \\
                                & \textbf{\textit{nand3}}          & 7815.93 \scriptsize{\textbf{(97.47\%)}}                     & \textbf{197.3073} \\
                                & \textbf{\textit{decoder2\_4}}    & 2324.89  \scriptsize{\textbf{(78.97\%)}}                    & \textbf{488.8251} \\
                                & \textbf{\textit{deco1m}}         & 8576.82  \scriptsize{\textbf{(97.14\%)}}                     & \textbf{244.7001} \\ \hline
\end{tabular}
\caption{Comparison of the time efficiency with individual embedding generation for each K-hop subgraph.}
\label{table:w/o our subgraph embedding extraction}
\vspace{-5mm}
\end{table}

In this section, we generate individual embeddings for each $K$-hop subgraph by applying the GNN separately to each K-hop graph, instead of following the approach described in Sec~\ref{2.3.2}, which extracts $K$-hop subgraph embeddings from the node embeddings of the entire graph.
Table ~\ref{table:w/o our subgraph embedding extraction} compares the running time between this method and our proposed approach for matching all target circuits.
We have the following observations: 
\textbf{1)} Our method achieves improved time efficiency by reducing running time across all target circuits when compared to \textbf{w/o our subgraph embedding extraction}.
This improvement stems from applying GNN once to the entire graph, while the \textbf{w/o our subgraph embedding extraction} approach requires GNN application for every $K$-hop graph.
\textbf{2)} We provide the reduction rate of our method relative to \textbf{w/o our subgraph embedding extraction} next to the running time. 
Our approach achieves up to a 97.47\% reduction in running time, indicating that it can achieve significantly more efficient matching in large-scale circuits.

\smallskip
\noindent\textbf{2) Negative Samples}

In this section, we analyze the influence of negative samples on matching performance. For the experiments, we utilize \textbf{\textit{col\_sel9\_512}} and \textbf{\textit{deco9\_512}} as the entire circuits and \textbf{\textit{nand3}} and \textbf{\textit{col\_sel1m}} as the target circuits.
We report the matching results for various combinations of positive samples (Pos) and four negative samples, including Partial (P), Mutation (M), Others (O), and Random (R).
The experiments are conducted under the five cases: 
1) Pos + R, 2) Pos + R + P, 3) Pos + R + M, 4) Pos + R + O, 5) ALL (Pos + R + P + M + O).

\begin{figure}[t] 
\begin{center}
\includegraphics[width=\linewidth]{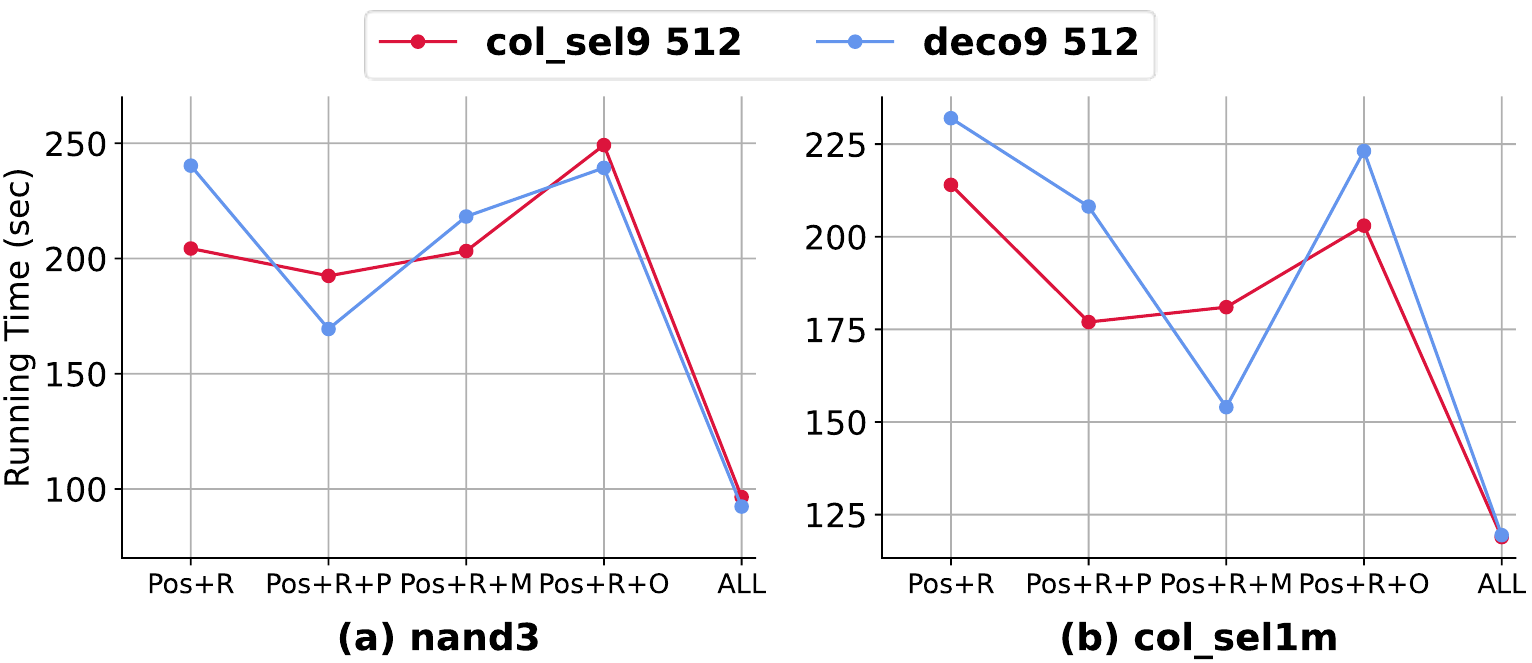}
\end{center}
\vspace{-1mm}
\caption{
Ablation studies on the impact of negative samples. }
\label{fig:ablation_studies}
\vspace{-5mm}
\end{figure}

Figure~\ref{fig:ablation_studies} shows the ablation studies regarding the impact of our proposed negative samples.
We have the following observations: \textbf{1)} We observe that \textbf{ALL}, which utilizes all four types of negative samples, achieves the lowest running time among the five cases. This indicates that using an appropriate combination of all four negative samples contributes to improved matching performance. This demonstrates that utilizing diverse negative samples improves the performance of target region prediction, ultimately resulting in more efficient matching.
\textbf{2)} Among the four types of negative samples, we observe that Partial and Mutation significantly contribute to reducing matching time. As hard negative samples, they enhance the accurate identification of target regions and enable more efficient target circuit matching.

\section{Conclusion} \label{conclusionsection}
In this work, we utilize GNN-based region prediction to identify regions with a high probability of containing the target circuit and perform target circuit matching based on these regions.
We explored an effective method for converting MOSFET-based netlists into graph representations and introduced four types of negative samples to enable efficient learning. 
 Additionally, to avoid the inefficiency of applying GNNs to all candidate subgraphs, we directly extract k-hop subgraph embeddings from the entire circuit using node embeddings from all hops.
 Our method demonstrates superior matching time efficiency compared to other models, driven by its effective target region prediction. Additionally, our method outperforms various baselines in target region prediction. Consequently, our model demonstrates greater applicability for subgraph matching in large-scale circuits compared to existing approaches. 
We expect that this efficient target graph matching method will play a key role in EDA and circuit analysis.

\section*{Acknowledgment} 
This work was supported by SK hynix Inc, the National Research Foundation of Korea(NRF) grant funded by the Korea government(MSIT) (RS-2024-00335098), and National Research Foundation of Korea(NRF) funded by Ministry of Science and ICT (RS-2022-NR068758).






\bibliographystyle{IEEEtran}
\bibliography{IEEEfull}

\end{document}